\documentclass[letterpaper, 10 pt, conference]{format/ieeeconf}

\usepackage{amsmath}
\makeatletter
\def\maketag@@@#1{\hbox{\m@th\normalfont\normalsize#1}}
\makeatother
\usepackage{amsfonts}

\usepackage{amsthm}
\usepackage[ruled,vlined]{algorithm2e}
\usepackage{mathtools}
\usepackage{tabularx}
\usepackage{graphicx}
\usepackage{subfigure}
\usepackage{enumerate}
\usepackage{float}
\usepackage{url}
\usepackage{verbatim}
\usepackage[linkcolor=black,citecolor=black,urlcolor=black,colorlinks=true]{hyperref}
\usepackage{cite}
\usepackage{hyperref}
\usepackage{algorithmicx}
\usepackage{array}
\usepackage{multirow}
\usepackage{longtable}
\usepackage{rotating}

\graphicspath{{figures/}}
\IEEEoverridecommandlockouts
\title{A Decoupled and Linear Framework for Global Outlier Rejection \\ over Planar Pose Graph}
\author{Tianyue Wu and Fei Gao\thanks{
		This work was supported by the National Natural Science Foundation of China under grant no. 62003299 and 62088101, and the Fundamental Research Funds for the Central Universities.}
	\thanks{ All authors are with the State Key Laboratory of Industrial Control Technology, Institute of Cyber-Systems and Control, Zhejiang University, Hangzhou, 310027, China.} \thanks{Email: {\tt\small \{modern\_gangster, fgaoaa\}@zju.edu.cn}}	\thanks{Corresponding Author: Fei Gao}}

\begin{document}

    \maketitle
    \thispagestyle{empty}
    \pagestyle{empty}

\begin{abstract}
We propose a robust framework for planar pose graph optimization contaminated by loop closure outliers. Our framework rejects outliers by first decoupling the robust PGO problem wrapped by a \emph{Truncated Least Squares} kernel into two subproblems. Then, the framework introduces a linear angle representation to rewrite the first subproblem that is originally formulated in rotation matrices. The framework is configured with the \emph{Graduated Non-Convexity} (\textmd{\texttt{GNC}}) algorithm to solve the two non-convex subproblems in succession without initial guesses. Thanks to the linearity property of the angle representation, our framework requires only a linear solver to optimally solve the optimization problems encountered in \textmd{\texttt{GNC}}. We extensively validate the proposed framework, named \textmd{\texttt{DEGNC-LAF}}~({\emph{DEcoupled Graduated Non-Convexity with Linear Angle Formulation}}) in planar PGO benchmarks. It turns out that it runs significantly (sometimes up to over 30 times) faster than the standard and \mbox{general-purpose} \textmd{\texttt{GNC}} while resulting in \mbox{high-quality} estimates.  

\end{abstract}
\vspace{-0.2cm}
\section{Introduction}
\vspace{-0.1cm}
\label{Sec1}
Pose Graph Optimization (PGO) is a fundamental estimation engine for Simultaneous Localization and Mapping~(SLAM) in many robotic applications. PGO aims to find poses of the robot that best explain the obtained noisy measurements. The solution space of the poses to be estimated in PGO lies in the special Euclidean group, which makes solving this problem untrivial.

Despite the challenges of PGO, recently proposed techniques (e.g.,~\texttt{SE-Sync}~\cite{rosen2019se}) can solve PGO optimally under mild preconditions. However, a portion of loop closure measurements are outliers in practice due to the inevitable perceptual aliasing. In the presence of outliers, even the optimal solution of PGO can still significantly deviate from the ground-truth. Therefore, it is common to use robust PGO \mbox{techniques\cite{sunderhauf2012switchable,olson2013inference,agarwal2013robust,latif2013robust,mangelson2018pairwise}} to gain resilience against outliers. 

However, the state-of-the-art methods for robust PGO are either \emph{local}~\cite{agarwal2013robust}, implying an over-dependence on a reliable initial guess, or consume heavy computational resources\mbox{\cite{tzoumas2019outlier,yang2020graduated}} and thus are far from being able to take on a broader range of tasks flexibly. Therefore, it is desirable to develop \emph{global}\footnote{We identify a method as global if it requires no initial guess or its result is independent of the initial guess.} outlier rejection techniques that often succeed in eliminating the misinformation  of outliers, while performing considerable efficiency.

In this paper, we show that we can fill the abovementioned gaps in the \emph{planar} PGO setup. Specifically, by decoupling rotation and translation estimates in planar PGO and adopting an angle setting for planar rotation to substitute rotation matrices, we construct two internally linear subproblems wrapped by the \emph{Truncated Least Squares} (TLS) kernel. We adopt the standard \emph{Graduated Non-Convexity}~(\texttt{GNC})\mbox{\cite{yang2020graduated,black1996unification}} algorithm to solve these two non-convex subproblems. Thanks to the linearity properties of the two subproblems, our proposed framework requires only linear solvers. 

\textbf{Contribution.} Our main contribution is to propose a novel framework to globally reject outliers over planar pose graph with considerable efficiency. The corresponding algorithm, named \emph{DEcoupled Graduated Non-Convexity with Linear Angle Formulation} (\texttt{DEGNC-LAF}), can (i) perform global outlier rejection while showing superior efficiency to \mbox{general-purpose} global methods (e.g.,~\cite{tzoumas2019outlier},~\cite{yang2020graduated}), and (ii) withstand outlier rates that may occur in real-world tasks and thus results in high-quality estimates. 

In addition, we also propose and validate that the first subproblem in a linear angle setting converges \texttt{GNC} with better accuracy and robustness (and much faster) than the one based on rotation matrix, which results in a significant performance enhancement of the complete estimation algorithm (see \emph{Remark 1} and Section~\ref{Sec7}).

We extensively validate the proposed approach in standard PGO benchmarks. The result illustrates that \texttt{DEGNC-LAF} is significantly faster than the standard and general-purpose \texttt{GNC} algorithm~\cite{yang2020graduated} while usually outperforming or matching \texttt{GNC} and local robust PGO techniques~\cite{agarwal2013robust,latif2013robust} in terms of accuracy and robustness.    

\section{Related Work}
\label{Sec2}
Ideas for designing robust PGO algorithms can be divided into two main categories: (i) reformulating to robustify the original PGO problem and solving the reformulated problem, and (ii) mining empirical or statistical evidence of outliers and filtering them out accordingly. 

\subsection{Reformulating to Gain Robustness against Outliers}
The existing robust PGO algorithms reformulating to robustify the original problem mainly follow three ways: (i) introducing \emph{switchable variables}, (ii) adopting the \emph{Consensus Maximization} paradigm~\cite{chin2017maximum,sarlette2009consensus} and (iii) adopting the \mbox{\emph{M-estimation}} paradigm~\cite{huber2011robust,bosse2016robust}.

\subsubsection{Switchable variable} The formulations proposed in~\mbox{\cite{sunderhauf2012switchable,agarwal2013robust,olson2013inference,doherty2022discrete}} introduce switchable variables for untrusted edges in the pose graph. For continuous optimization problems with switchable variables~(e.g.,\mbox{\cite{sunderhauf2012switchable,agarwal2013robust,olson2013inference}}), standard nonlinear optimization techniques can be used to solve them. Some off-the-shelf libraries, such as ceres-solver~\cite{Agarwal_Ceres_Solver_2022} and g2o~\cite{kummerle2011g}, provide implementations of these techniques. By contrast, the switchable variables in~\cite{doherty2022discrete} are binary (i.e., discrete), so an alternating~minimization framework proposed by the authors was used to solve the problem. Unfortunately, the solving techniques mentioned above are all local, requiring reliable initial guesses.

\subsubsection{Consensus Maximization} Consensus Maximization (CM) looks for an estimate that maximizes the number of measurements with errors under a prescribed threshold. \texttt{RANSAC} \cite{hartley2003multiple} is a popular heuristic for CM and requires no initial guess, but the performance of \texttt{RANSAC} is indeterministic and its runtime grows exponentially with increasing outlier rates. Tzoumas et~al.~\cite{tzoumas2019outlier} proposed \texttt{ADAPT}, a \mbox{general-purpose} heuristic algorithm to solve \emph{Minimally Trimmed Squares} (MTS) estimation which shares commonalities with CM \cite{antonante2021outlier}.

\subsubsection{M-estimation} M-estimation robustifies the original cost function of a problem via wrapping it in a robust kernel. Local nonlinear optimization is an efficient way to solve robust structure from motion~(SFM)~\cite{chatterjee2017robust, schonberger2016structure} and robust PGO~\cite{casafranca2013back} formulated as M-estimation problems. To get rid of the dependence of robust \mbox{kernel-wrapped} PGO problems on reliable initial guesses, Carlone and Calafiore~\cite{carlone2018convex} proposed convex relaxations for the $l_{1}$-norm and Huber kernels. \emph{Graduated Non-Convexity} (\texttt{GNC}) has been used in robust estimation problems~\cite{black1996unification,zhou2016fast} as a heuristic to globally solve the \mbox{M-estimation} problem. More recently, Yang \emph{et~al.}~\cite{yang2020graduated} combined modern global non-minimal solvers (e.g., \texttt{SE-Sync}) with \mbox{Black-Rangarajan} duality~\cite{black1996unification} for \mbox{M-estimation}, which benefits our work (see Section \ref{Sec5}).

\subsection{Other Models/Evidence for Outliers}
Latif \emph{et~al.}~\cite{latif2013robust} proposed the \emph{Realizing, Reversing, and Recovering} (\texttt{RRR}) algorithm to reject outliers based on the residuals of local PGO. Mangelson \emph{et~al.}~\cite{mangelson2018pairwise} proposed \emph{Pairwise Consistency Maximization}~(\texttt{PCM}) to select a set of inliers, which checks pairwise consistency for each two loop closures and infers the set of inliers from a graphical representation of pairwise consistency. Indelman~\emph{et~al.}~\cite{indelman2016incremental} modeled the problem of finding loop closure outliers as a \emph{maximum a posteriori estimation} problem. The authors inferred the formulated problem via \emph{expectation maximization}, to which a similar idea can be found in~\cite{karimian2020rotational}.

\section{Decoupled Robust Planar PGO}
\label{Sec3}
\subsection{Coupled and Rotation Matrix-based Planar PGO}
\label{Sec3A}
Planar PGO solves 2D poses $(R_i,t_i)$ sampled along the robot trajectory, where $R_i \in \mathrm{SO}(2) \text { and } t_i \in \mathbb{R}^{2}$. Typically, we anchor the first pose and assume that the noisy measurements $(\tilde{R}_{ij},\tilde{t}_{ij})$ obtained by the robot are sampled from the following probabilistic generative model\cite{rosen2019se}: 
\begin{subequations}
	\begin{equation}
		\label{eq:RotationMeasurement}
		\ \ \ \ \tilde{R}_{ij}=R_{ij}R_{ij}^{\epsilon},\qquad  R_{ij}^{\epsilon}\sim Langevin\left( I_d,\kappa _{ij} \right),
		\vspace{0.01cm} 
	\end{equation}
	\begin{equation}
		\label{eq:TranslationMeasurement}
		\tilde{t}_{ij}=t_{ij}+t_{ij}^{\epsilon},\mathrm{          }t_{ij}^{\epsilon}\sim \mathcal{N} \left( 0,\tau _{ij}^{-1}I_d \right), 
	\end{equation}
\end{subequations}
where for all pairwise poses $\left( {i,j} \right)$, $\left( {t_{ij},R_{ij}} \right)$ is the true but latent value of relative poses. 

Given the noisy measurements and substituted into the probability model of Langevin~\cite{boumal2014cramer} and Gaussian distribution appearing in (1a) and (1b), the (coupled and rotation \mbox{matrix-based}) PGO is modeled as a \emph{maximum likelihood estimation} problem:
\begin{equation}
	\label{eq:PGO}
	(\boldsymbol{\hat{R}},\boldsymbol{\hat{t}})\underset{\footnotesize\begin{matrix}
			{\ R_{i} \in {\rm SO}{(2)}} \\
			{\ t_{i} \in \mathbb{R}^{2}} \\
	\end{matrix}}{= {\rm arg\, min}}{\sum\limits_{({i,j})}\left\{ \begin{array}{l}
			{\kappa_{ij}{\parallel {R_{j} - R_{i}{\tilde{R}}_{ij}} \parallel}_{F}^{2}} \\
			{+ \tau_{ij}{\parallel {t_{j} - t_{i} - R_{i}{\tilde{t}}_{ij}} \parallel}_{2}^{2}} \\
		\end{array} \right\}},
\end{equation}
where $(\boldsymbol{\hat{R}},\boldsymbol{\hat{t}})$ is a compact representation of the solution to problem (\ref{eq:PGO}). 

Problem (2) is currently proposed to be tightly relaxed as a (convex) \emph{semidefinite programming} under mild preconditions~\cite{rosen2019se}, and thus can be solved optimally. However, in real-world tasks, a fraction of the loop closure measurements are outliers. The original formulation (2) is sensitive to these outliers, so it is common to apply robust kernels to wrap the original problem to gain resilience against the outliers.

\subsection{Robust Planar PGO with Truncated Least Squares Kernel}
\label{Sec3B}
To limit as much as possible the sensitivity of planar PGO to outliers, we adopt a \mbox{TLS-based} scheme to reformulate the planar PGO problem (2), which leads to the \emph{Coupled and Rotation Matrix-Based Truncated Least Squares Planar PGO (\mbox{TLS-PGO})} problem:
\begin{equation}
	\hspace{-2.2mm}\min_{\footnotesize\begin{array}{c}
			t_i\in \mathbb{R} ^2\\
			R_i\in \mathrm{SO(}2)\\
	\end{array}}\hspace{-2mm} {\sum_{\left( i,j \right) \in \mathcal{E} _{od}}{\hspace{-2.2mm}{\left( r_{ij}^{pgo} \right) ^2}}+\sum_{\left( i,j \right) \in \mathcal{E} _{lc}}{\hspace{-2.2mm}{\rho _c\left( r_{ij}^{pgo} \right)}}},
\end{equation}
where
\begin{equation}
\rho _c\left( r \right) =\min \left( r^2,c^2 \right) 
\end{equation}
is the TLS kernel, $\mathcal{E} _{od}$ and $\mathcal{E} _{lc}$ are the sets consisting of odometry and loop closure respectively, and
\begin{equation}
r_{ij}^{pgo}{= \sqrt{\kappa_{ij}{\parallel {R_{j} - R_{i}{\tilde{R}}_{ij}} \parallel}_{F}^{2} + \tau_{ij}{\parallel {t_{j} - t_{i} - R_{i}{\tilde{t}}_{ij}} \parallel}_{2}^{2}}
}.
\end{equation}

In problem (3), we only apply robust kernels to loop closure measurements, since we can typically trust odometry measurements as inliers.

\subsection{Decoupled Robust Planar PGO}
In this subsection, we propose a decoupled formulation for robust planar PGO. We decouple problem (3) into two subproblems that are solved successively. The subproblem to be solved first is the \emph{Truncated Least Squares Rotation Matrix-Based Planar Rotation Averaging (TLS-RA)} problem:
\begin{equation}
	\hspace{-2.2mm}\min_{\footnotesize\begin{array}{c}
			R_i\in \mathrm{SO(}2)\\
	\end{array}}\hspace{-2mm} {\sum_{\left( i,j \right) \in \mathcal{E} _{od}}{\hspace{-2.2mm}{\left( r_{ij}^{ra} \right) ^2}}+\sum_{\left( i,j \right) \in \mathcal{E} _{lc}}{\hspace{-2.2mm}{\rho _{c_1}\left( r_{ij}^{ra} \right)}}},
\end{equation}
where
\begin{equation}
	r_{ij}^{ra}{= \sqrt{\kappa_{ij}{\parallel {R_{j} - R_{i}{\tilde{R}}_{ij}} \parallel}_{F}^{2}}
	}.
\end{equation}

Once problem (5) is solved, the rotation $R_i$ will be fixed in the second subproblem. Thereby, we have the second subproblem, the \emph{Truncated Least Squares Rotation Matrix-Based Planar Translation Averaging (TLS-TA)} problem:
\begin{equation}
	\hspace{-2.2mm}\min_{\footnotesize\begin{array}{c}
			t_i\in \mathbb{R} ^2\\
	\end{array}}\hspace{-2mm} {\sum_{\left( i,j \right) \in \mathcal{E} _{od}}{\hspace{-2.2mm}{\left( r_{ij}^{ta} \right) ^2}}+\sum_{\left( i,j \right) \in \mathcal{E} _{lc}}{\hspace{-2.2mm}{\rho _{c_2}\left( r_{ij}^{ta} \right)}}},
\end{equation}
where
\begin{equation}
	r_{ij}^{ta}{= \sqrt{\tau_{ij}{\parallel {t_{j} - t_{i} - \hat{R_{i}}{\tilde{t}}_{ij}} \parallel}_{2}^{2}}
	},
\end{equation}
where $\hat{R}_{i}$ is the rotation estimate determined by solving problem (6) and remains constant in problem (8).

We note that the estimate obtained by successively optimally solving subproblem (6) and subproblem (8) is not equivalent to the one obtained by optimally solving problem~(3). But an important fact is that empirically, the optimal rotation estimate from solving problem~(6) is close to the one from solving problem (3). This can be explained by the fact that (i) in the absence of outliers, these two solutions are usually close to each other, as shown in Table 1, and (ii) if problems~(3) and (6) are solved optimally in the presence of outliers, the solutions will be close to those in the absence of outliers due to the robustness of TLS kernel, and thus are close to each other according to (i). Therefore, robust rotation estimation does not undergo significant distortion due to decoupling. This implies the fact that translation estimates obtained from optimally solving problem~(8) are consistent with those obtained from optimally solving problem~(3).

\section{Robust Planar Rotation Estimation with Angle-Based Linear Formulation}
\label{Sec4}
Although rotation matrices can extensively represent planar and spatial rotations without singularity, the planar rotation has another direct \emph{angle} representation that is less mentioned today but enables the decoupled framework to work better. In this section we show how to use this angle setting to construct a problem to replace the problem (6) in the decoupled framework.
\vspace{-0.1cm}
\subsection{Linear Anglular Domain for Planar Rotation}
\vspace{-0.1cm}
\label{Sec4A}

We use the robot's orientation angle $\theta_{i}$ instead of the rotation matrix to represent the rotational component of the robot's pose, so that the probabilistic generative model of the rotation measurements is replaced from Eq. (1a) with the following model:
\begin{equation}
\begin{split}
\tilde{\theta}_{ij}=\left< \left. \theta _j-\theta _i+\theta _{ij}^{\epsilon} \right> \right. =\theta _j-\theta _i+k_{ij}2\pi +\theta _{ij}^{\epsilon}, \\
\theta _{ij}^{\epsilon}\sim \mathcal{N} \left( 0,\kappa _{ij}^{-1}I_d \right), 
\end{split}
\end{equation}
where $\langle\cdot\rangle: \mathbb{R} \rightarrow(-\pi,+\pi]$ is the 2D \emph{geodesic distance} and $k_{i j} \in \mathbb{Z}$ is called a \emph{regularization variable} that regularizes the angle measurements in the interval $(-\pi,+\pi]$. 

We now consider the rotation averaging problem in this orientation angle setting when no robust kernel is applied:  
\begin{equation}
	\min_{\footnotesize\begin{matrix}
			{\theta _{i} \in \mathbb{R}}\vspace{-0.5ex} \\ 
		{\hspace{-1.2mm}{\ k_{i j} \in \mathbb{Z}}} \\ 
	\end{matrix}}{\sum\limits_{({i,j})} \begin{array}{l}
			{\kappa_{ij}{\parallel {\theta _j - \theta _i + k_{ij}2\pi - \tilde{\theta}_{ij}} \parallel}_{2}^{2}} 
		\end{array}}.
\end{equation}
We can clearly see that if the regularization variable~$k_{ij}$ can be determined as a priori and fixed, we can obtain a \emph{linear estimation} problem from (11), which will make its robust version much less difficult to solve (see \emph{Remark 1}).

We adopt the approximate method proposed in \cite{carlone2014fast} to estimate the regularization variable $k_{ij}$ without solving the integer-mixed problem (11). $k_{ij}$ can be determined based on the fact that in the absence of noise, the measurements~$\tilde{\theta}_{ij}$ along each \emph{cycle} in the pose graph have to sum-up to zero, of which a sufficient and necessary condition is that the measurements along each cycle in the \emph{cycle basis}~\cite{bai2021sparse} of the pose graph sum-up to zero. According to this fact, the integer $k_{ij}$ can be closed-form solved from a linear system. In the presence of noise, $k_{ij}$ solved from the linear system will no longer be an integer, but the corresponding regularization variable can be rounded to the integer closest to it. This rounding scheme is perfectly accurate in the majority of real-world scenarios~\cite{carlone2014fast}, so we can convincedly adopt it to determine $k_{ij}$ as a priori, thus converting problem~(10) into a linear estimation.

\begin{table}[]
	\caption{Average Rotation Error (ARE) between Optimal Solutions to PGO and Rotation Averaing on Standard Datasets}
	\label{table1}
	\setlength{\tabcolsep}{26pt}
	\renewcommand\arraystretch{1.5}
	\centering
	\begin{tabular}{cc}
		\hline
		\textbf{Dataset (no outliers)}   & \textbf{ARE (deg)} \\
		\hline
		city10000 & 0.6971    \\
		intel     & 0.9215    \\
		kitti\_00 & 0.6453     \\
		kitti\_02 & 0.3268    \\
		kitti\_05 & 0.1494    \\
		manhattan & 0.5306    \\
		csail     & 0.0631    \\ 
		\hline
	\end{tabular}
	\vspace{-0.5cm}
\end{table}
\vspace{-0.1cm}
\subsection{Robust Rotation Averaging in Linear Angle Setting}
\vspace{-0.1cm}
\label{Sec4B}
By adopting the linear angle setting shown in (11), we can replace problem (6) with the \emph{Truncated Least Squares Angle-Based Planar Rotation Averaging (TLS-ARA)} problem:
\begin{equation}
	\hspace{-2.2mm}\min_{\footnotesize\begin{array}{c}
			\theta _{i} \in \mathbb{R}\\
	\end{array}}\hspace{-2mm} {\sum_{\left( i,j \right) \in \mathcal{E} _{od}}{\hspace{-2.2mm}{\left( r_{ij}^{ara} \right) ^2}}+\sum_{\left( i,j \right) \in \mathcal{E} _{lc}}{\hspace{-2.2mm}{\rho _{c_1}\left( r_{ij}^{ara} \right)}}},
\end{equation}
where
\begin{equation}
	r_{ij}^{ara}{=\sqrt{\kappa_{ij}{\parallel {\theta _j - \theta _i + k_{ij}2\pi - \tilde{\theta}_{ij}} \parallel}_{2}^{2}}}.
\end{equation}

Although both problem (12) and problem (6) are \mbox{non-convex} and difficult to solve directly, intuitively, problem~(12) will be easier to solve because it is a robust \emph{linear} estimation problem, whereas problem~(6) is still a nonlinear problem in the absence of the robust kernel.

The following section demostrates how to solve problems~(3), (6), (8), (12) using a general \texttt{GNC} framework.

\section{Solving Robust PGO via Graduated Non-Convexity}
\label{Sec5}
Firstly, we note that problems (3), (6), (8), (12) share the uniform form:
\begin{equation}
	\hspace{-2.2mm} {\rm min}\hspace{-2mm} {\sum_{\left( i,j \right) \in \mathcal{E} _{od}}{\hspace{-2.2mm}{\left( r_{ij} \right) ^2}}+\sum_{\left( i,j \right) \in \mathcal{E} _{lc}}{\hspace{-2.2mm}{\rho _{c}\left( r_{ij} \right)}}},
	\vspace{0.01cm}
\end{equation}
which can be equivalently rewritten as follows:
\begin{equation}
	\small\hspace{-2.2mm} {\rm min} \left\{\hspace{-2mm} {\hspace{1mm}\sum_{\left( i,j \right) \in \mathcal{E} _{od}}{\hspace{-2.2mm}{\left( r_{ij} \right) ^2}}+\sum_{\left( i,j \right) \in \mathcal{E} _{lc}}{\hspace{-2.2mm}\underset{w _{ij}\in \left\{ 0,1 \right\}}{\min}\left[ w_{ij}r_{ij}^{2}+\left( 1-w_{ij} \right) c^2 \right] }}\right\}.
\end{equation}

Problem (15) allows the use of (i) \emph{Alternating Optimizing}~(\texttt{AM}), (ii) \emph{Semidefinite Programming} (\texttt{SDP})~\cite{yang2022certifiably} and (iii) \emph{Graduated Non-Convexity} (\texttt{GNC})~\cite{yang2020graduated} to solve it. However, while \texttt{AM} is efficient, it requires an initial guess and can easily fall into local minima. By contrast, \texttt{SDP} can provide optimality certificates, but is very slow to solve at this stage. Therefore, we adopt \texttt{GNC}, which requires no initial guesses and can effectively avoid local minima while performing good efficiency, to solve the problem.

Applying a \emph{GNC function}~\cite{black1996unification} controlled by a parameter $\mu$ to approximate the TLS kernel, \texttt{GNC} uses a \emph{continuation}-type algorithm~\cite{carlone2022estimation} to optimize the following problem instead of problem (15):
\begin{equation}
	\small\min \hspace{-1mm} \left\{\hspace{-1mm} \sum_{\left( i,j \right) \in \mathcal{E} _{od}}{\hspace{-2.2mm}{\left( r_{ij} \right) ^2}} +\hspace{-1mm} \hspace{-2mm}
	 \sum_{\left( i,j \right) \in \mathcal{E} _{lc}} {\hspace{-2mm}\min _{w_{ij} \in[0,1]}\left[w_{ij} r_{ij}^{2}+\frac{\mu\left(1-w_{ij}\right)}{\mu+w_{ij}} c^{2}\right]}\hspace{-1mm}\right\},
\end{equation}
for which the parameter $\mu$ is gradually updated by multiplying a continuation factor~$f$ at a time. Intuitively, updating $\mu$ implies adjusting the convexity of the kernel that will gradually revert to TLS as $\mu$ increases.

\texttt{GNC} is initialized by solving a globally solvable problem, i.e., the optimization problem in the form of (16) with $\mu$ tending to zero and all $w_{ij}$ being~1, shown as follows:
\begin{equation}
	\min \sum_{\left( i,j \right) }{\left( r_{ij} \right) ^2},
\end{equation}
which can be solved optimally with global PGO techniques~\cite{rosen2019se} with $r_{ij}$ specified as $r_{ij}^{pgo}$ (5), global rotation averaing techniques~\cite{rosen2019se,parra2021rotation} with $r_{ij}$ specified as $r_{ij}^{ra}$ (7) and linear solver with $r_{ij}$ specified as $r_{ij}^{ta}$ (9) and $r_{ij}^{ara}$ (11). 

The solution to problem (17) will be used as the initial guess for subsequent optimizations in \texttt{GNC}. For each update of $\mu$ (called one iteration), \texttt{GNC} alternately optimizes two subproblems of problem (15): the first subproblem optimizes $w_{ij}$, where $r_{ij}$ is determined by the solution of the second subproblem in the previous iteration, for which a closed-form solution exists~(cf.~\cite{yang2020graduated}); the second subproblem optimizes $r_{ij}$, where $w_{ij}$ is determined by the first subproblem in the current iteration. The second subproblem can be considered a weighted version of problem~(17) and thus can be solved globally with the same tools. \texttt{GNC} will keep iterating until all $w_{ij}$ become binaries. We can identify a measurement as an outlier if its corresponding $w_{ij}$ converges to 0.

\section{Decoupled Graduated Non-Convexity with Linear Angle Formulation}
\label{Sec6}
By introducing the angle setting and \texttt{GNC} algorithm, we proposed a decoupled and linear framework, named \emph{DEcoupled Graduated Non-Convexity with Linear Angle Formulation}~(\texttt{DEGNC-LAF}), to globally reject outliers over planar pose graph, the pipeline of which is shown as follows:
\begin{itemize}
\item [1)] Computing regularization variables $k_{ij}$ in Eq. (13) 
\item [2)] Using \texttt{GNC} with $k_{ij}$ to solve TLS-ARA (problem~(12)) to obtain the rotation estimate $\hat R_i$ 
\item [3)] Using \texttt{GNC} with $\hat R_i$ to solve TLS-TA (problem~(8)) and rejecting outliers according to the resulting $w_{ij}$ ($w_{ij}$ corresponding to outliers will converges to zero) 
\item [4)] Using \texttt{SE-Sync} with inliers to finally estimate rotations and translations $(\boldsymbol{\hat{R}},\boldsymbol{\hat{t}})$   
\end{itemize}
The pseudocode is summarized in Algorithm 1.

We note that instead of taking the solutions to TLS-ARA and TLS-TA as the output estimates, we use the rotation estimates of TLS-ARA to serve for solving TLS-TA and reject outliers based on the resulting $w_{ij}$ of TLS-TA. We end up optimizing poses in coupled PGO using \texttt{SE-Sync}. Such an approach will result in more accurate estimates. 

\emph{Remark 1 (Benefits of Decoupling and Linear Formulation):} Linear formulation (9) and (13) enables the initialization problem (17) and the second subproblem of (15) in the alternating optimization of \texttt{GNC} linearly solvable, thus ensuring that optimal solutions can be obtained. In contrast, in the iterations of solving the problem~(3) and problem (7), \texttt{GNC} needs to repeatedly call nonlinear global solvers which require preconditions to get the global optimal solution. As a result, the solvers may fail to get the optimal solution in the first few iterations, which increases the risk of \texttt{GNC} falling to a local minima. In addition, linear systems can be solved exceptionally efficiently and do not require the computation of optimality certificates which is rather time-consuming. 
\vspace{-1mm}
\begin{algorithm}
	\caption{\emph{DEcoupled Graduated Non-Convexity with Linear Angle Formulation} (\texttt{DEGNC-LAF})}
	\KwIn{reduced incidence matrix $\boldsymbol{A}$ (topology of the pose graph), measurements set $\{(\tilde{\theta}_{ij},\tilde{t}_{ij})\}$, noise level sets $\{\kappa _{ij}\}$ and $\{\tau_{ij}\}$, odometry~set~$\mathcal{E} _{od}$, loop closure set~$\mathcal{E} _{lc}$, threshold~$c_1$~(default: \texttt{chi2inv}(0.99, 1)) threshold~$c_2$~(default: \texttt{chi2inv}(0.99, 2)), and \texttt{GNC}~continuation factor $f$ (default: 1.4);}
	\KwOut{inlier set $\mathcal{I}$, poses estimate $(\boldsymbol{\hat{R}},\boldsymbol{\hat{t}})$}
	\Begin
	{
		// \textsf{\smaller Compute regularization variables } \\ 
		$\{k_{ij}\}\gets$ {\texttt{Regularize}}($(\tilde{R}_{ij},\tilde{t}_{ij})$,~$\boldsymbol{A}$);\\	
		// \textsf{\smaller Solve TLS-ARA using \texttt{GNC}}  \\
		$\{\hat \theta_i\}\gets$ \texttt{GNCforARA}($\{\tilde{\theta}_{ij}\}$,~$\{\kappa _{ij}\}$,~$\mathcal{E} _{od}$,~$\mathcal{E} _{lc}$,~$c_1$,~$f$,~$\boldsymbol{A}$); \\
		// \textsf{\smaller Convert angle to rotation matrix} \\
		$\{\hat R_i\}\gets$ \texttt{RecoverR}($\{\hat \theta_i\}$); \\
		// \textsf{\smaller Solve TLS-TA using \texttt{GNC}} \\
		$\mathcal{I}\gets$ \texttt{GNCforTA}($\{\tilde{t}_{ij}\}$,~$\{\hat R_i\}$,~$\{\tau _{ij}\}$,~$\mathcal{E} _{od}$,~$\mathcal{E} _{lc}$,~$c_2$,~$f$,~$\boldsymbol{A}$); \\
		$\{\tilde R_{ij}\}\gets$ \texttt{RecoverR}($\{\tilde \theta_{ij}\}$); \\
		//  \textsf{\smaller Final estimate} \\
		$(\boldsymbol{\hat{R}},\boldsymbol{\hat{t}})\gets$ \texttt{SE-Sync}($\{(\tilde{R}_{ij},\tilde{t}_{ij})\}$,~$\{\kappa _{ij}\}$,~$\{\tau _{ij}\}$,~$\boldsymbol{A}$); \\	
		\Return{$\mathcal{I}$, $(\boldsymbol{\hat{R}},\boldsymbol{\hat{t}})$};
	}
\end{algorithm}
\vspace{-4mm}

\begin{figure*}
	\centering
	\includegraphics[width=2.0\columnwidth]{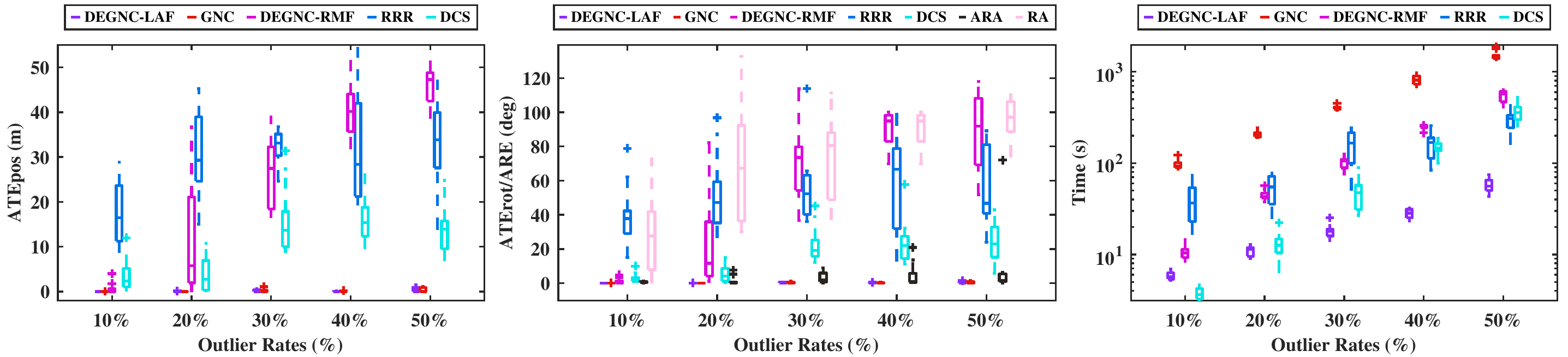}
	\vspace{-0.3cm}
	\caption{\textbf{city10000} \textbf{(}5000 poses\textbf{)}\textbf{.} Average Trajectory Error (ATE), Average Rotation Error (ARE) and the running time of the proposed approach compared to state-of-the-art techniques on the \textsf{\smaller city10000} dataset. Statistics are computed over 10 Monte Carlo runs. 
		\label{Fig_city10000}}
	\vspace{-0.3cm}
\end{figure*}
\begin{figure*}
	\centering
	\includegraphics[width=2.0\columnwidth]{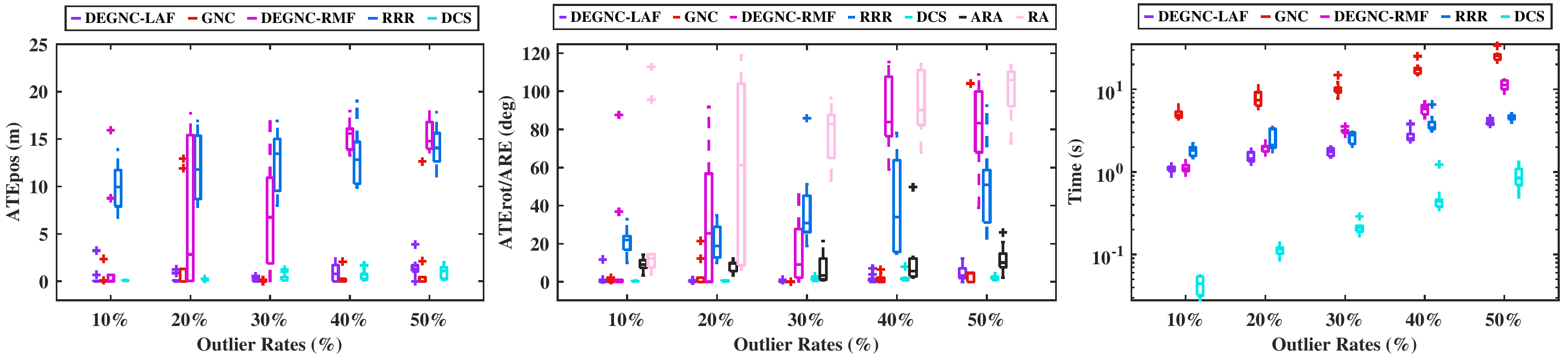}
	\vspace{-0.3cm}
	\caption{\textbf{intel}\textbf{.} Average Trajectory Error (ATE), Average Rotation Error (ARE) and the running time of the proposed approach compared to state-of-the-art techniques on the \textsf{\smaller intel} dataset. Statistics are computed over 10 Monte Carlo runs.
		\label{Fig_intel}}
	\vspace{-0.3cm}
\end{figure*}
\begin{figure*}
	\centering
	\includegraphics[width=2.0\columnwidth]{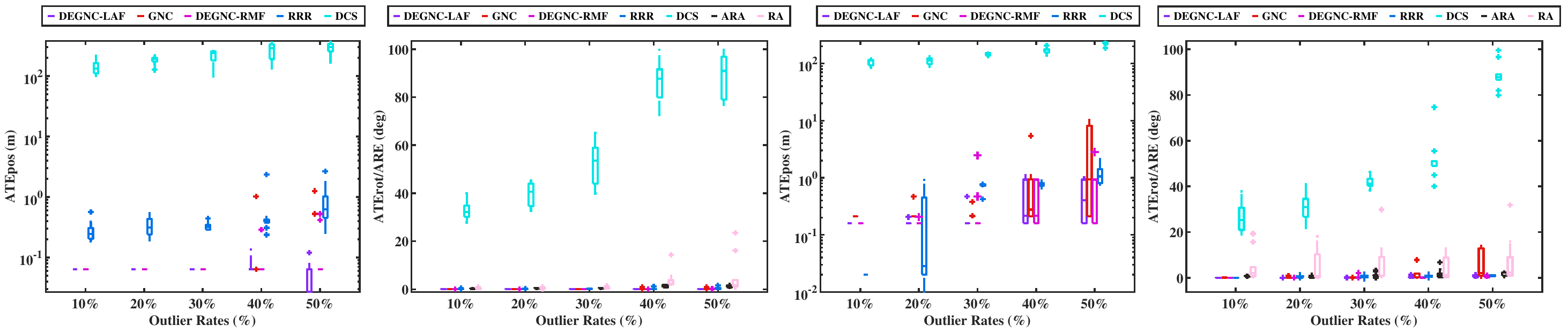}
	\vspace{-0.3cm}
	\caption{\textbf{kitti\_00 and kitti\_05}\textbf{.} Average Trajectory Error (ATE) and Average Rotation Error (ARE) of the proposed approach compared to state-of-the-art techniques on the \textsf{\smaller kitti\_00} dataset (first to second from left) and the \textsf{\smaller kitti\_05} dataset (third to fourth from the left). Statistics are computed over 10 Monte Carlo runs.
		\label{Fig_kitti}}
	\vspace{-0.5cm}
\end{figure*}

\begin{table}[]
	\caption{Information of the Datasets Tested on}
	\label{table2}
	\setlength{\tabcolsep}{24pt}
	\renewcommand\arraystretch{1.5}
	\centering
	\begin{tabular}{ccc}
		\hline
		\textbf{Dataset}   & \textbf{Poses} & \textbf{Loop Closures} \\ 
		\hline
		city10000 & 5000 & 3385     \\
		intel     & 1728 & 786    \\
		kitti\_00 & 4541 & 137     \\
		kitti\_05 & 2761 & 66    \\
		manhattan & 3500 & 1953    \\
		\hline
	\end{tabular}
	\vspace{-0.5cm}
\end{table}

\section{Experiments}
\label{Sec7}
\subsection{Experimental Setup}
\label{Sec7A}
	We test the performance of \texttt{DEGNC-LAF} on five standard PGO benchmarking datasets: \textsf{\small city10000}, \textsf{\small intel}, \textsf{\small kitti\_00}, \textsf{\small kitti\_05} and \textsf{\small manhattan}, described in \cite{rosen2019se}. Some information of these datasets are listed in Table 2. We benchmark our approach against (i) \texttt{GNC}~\cite{yang2020graduated}, (ii) \emph{DEcoupled Graduated Non-Convexity with Rotation Matrix Formulation} (\texttt{DEGNC-RMF}), which is the rotation matrix version of \texttt{DEGNC-LAF} (i.e., the first subproblem in the proposed framework is replaced by TLS-ARA with TLS-RA), (iii) \emph{Realizing, Reversing, and Recovering}~(\texttt{RRR})~\cite{latif2013robust} and (iv) \emph{Dynamic Covariance Scaling}~(\texttt{DCS})~\cite{agarwal2013robust}. Since \texttt{GNC} takes too much time on the original \textsf{\small city10000} dataset, we select only the first 5,000 poses with corresponding measurements of the dataset. While the datasets we test on contain no outlier initially, we select random pairs of poses and add a loop closure between each pair of poses to simulate outliers. 

\vspace{-0.22mm}
We performed all experiments in C++ running on a Linux machine with the Intel i7-12700KF (3.60 GHz). We use the latest accelerated version~\cite{Rosen2022Accelerating} of \texttt{SE-Sync} and \texttt{SO-Sync} (i.e., the rotation averaging version of \texttt{SE-Sync}) to respectively configure \texttt{GNC} and \texttt{DEGNC-RMF} and use 4 threads to drive them, while \texttt{DEGNC-LAF} is implemented with linear solver used in \texttt{GTSAM}~\cite{gtsam} using only one thread. The relevant parameters of \texttt{GNC} follow the configuration in~\cite{yang2020graduated}. We selected an open source\footnote{\textsf{https://github.com/gisbi-kim/toy-robust-backend-slam}} version of \texttt{DCS} implemented with ceres-solver~\cite{Agarwal_Ceres_Solver_2022}. We use kernel size $\Phi$ = 1 for \texttt{DCS} and 4 threads to drive the ceres-solver, and set the maximum number of iterations to 100, keeping default settings  in the source code for all other parameters. We use the \texttt{RRR} algorithm implemented by the authors and set the clustering threshold $\gamma$ =~10.

\begin{figure*}
	\centering
	\includegraphics[width=2.0\columnwidth]{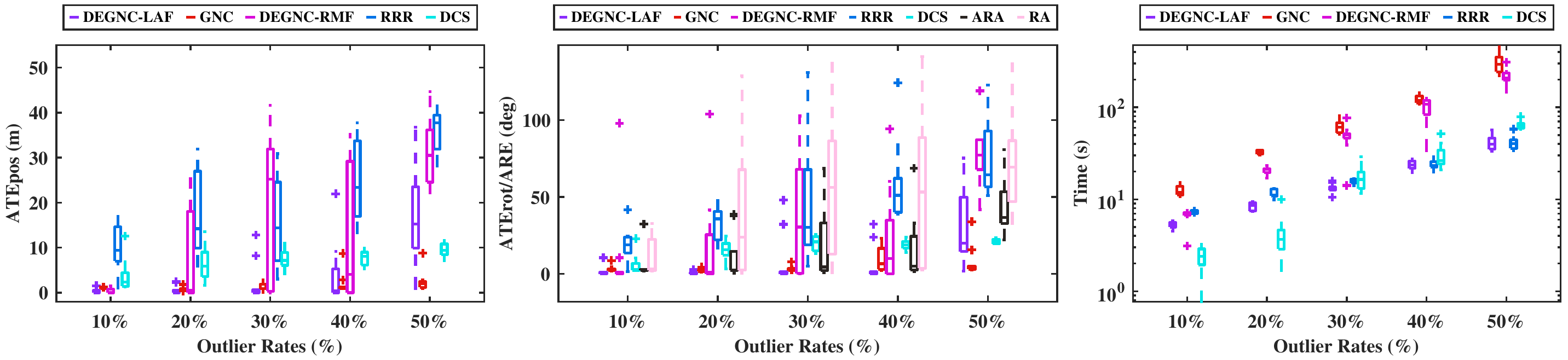}
	\vspace{-0.3cm}
	\caption{\textbf{manhattan}\textbf{.} Average Trajectory Error (ATE), Average Rotation Error (ARE) and the running time of the proposed approach compared to state-of-the-art techniques on the \textsf{\smaller manhattan} dataset. Statistics are computed over 10 Monte Carlo runs.
		\label{Fig_manhattan}}
	\vspace{-0.5cm}
\end{figure*} 

\subsection{Results}
\label{Sec7B}
We evaluate the algorithms by Average Trajectory Error (i.e., the $\mathrm{ATE}_{\mathrm{pos}}$ and $\mathrm{ATE}_{\mathrm{rot}}$ in \cite{Zhang18iros}) and running time. When evaluating $\mathrm{ATE}_{\mathrm{rot}}$ we also show Average Rotation Error~(ARE) of the solutions to TLS-ARA and TLS-RA with respect to optimal rotation estimates obtained by \texttt{SO-Sync} in the absence of outliers, which are labeled as "ARA" and "RA" respectively in the figures.  

\textbf{city10000}. Fig. 1 shows the performance of each algorithm on the \textsf{\small city10000} dataset. \texttt{DEGNC-LAF} and \texttt{GNC} dominate the other algorithms in terms of accuracy and robustness against outliers. At the same time, \texttt{DEGNC-LAF} is the fastest approach in most cases, except for being slightly slower than \texttt{DCS} at the outlier rate of 10\%. Although \texttt{GNC} can achieve similar accuracy to \texttt{DEGNC-LAF}, it is sometimes over 30 times slower than the latter, since it consumes too much time to compute optimality certificates.

\textbf{intel}. Fig. 2 shows the performance of each algorithm on the \textsf{\small intel} dataset. On this dataset, both \texttt{DEGNC-LAF} and \texttt{GNC} can achieve moderate robustness and accuracy. Despite this, \texttt{DEGNC-LAF} runs 4-7 times faster than \texttt{GNC}. This dataset allows \texttt{DCS} to maintain more stable performance compared to \texttt{DEGNC-LAF} and \texttt{GNC} while running at great speed. This is probably because the \textsf{\small intel} dataset provides a good initial guess for \texttt{DCS} so that it does not take a few iterations to reach the global minima. 

\textbf{kitti\_00 and kitti\_05}. These two datasets do not provide initial guesses, so we use the groud-truth as the initial guesses for \texttt{RRR} and \texttt{DCS}. Loop closures in these two datasets are very sparse, so each algorithm can perform good efficiency and thus we do not show the running time in Fig. 3. \texttt{DEGNC-LAF}, \texttt{GNC} and \texttt{DEGNC-RMF} perform good robustness and high accuracy on the \textsf{\small kitti\_00} dataset. In addition, the solution to \mbox{TLS-ARA} always shows small errors on this dataset. \texttt{DEGNC-LAF} outperforms other algorithms on the \textsf{\small kitti\_05} dataset, being robust to 50\% of outliers, while \texttt{GNC} begins to break at 40\% of outliers.  

\textbf{manhattan}. \textsf{\small manhattan} is a particularly challenging dataset for robust planar PGO. The odometry measurements in \textsf{\small manhattan} have large rotational errors, which causes \texttt{DEGNC-LAF} and \texttt{DEGNC-RMF} to break when solving \mbox{TLS-ARA} and TLS-RA in cases of high outlier rates~(i.e., 50\%), as shown in Fig. 4. Although \texttt{GNC} shows the best robustness on this dataset, it sometimes fails and takes hundreds of seconds at high outlier rates. At outlier rates below 30\%, \texttt{DEGNC-LAF} can achieve competitive accuracy with \texttt{GNC} and outperforms other algorithms, while runs 2-4 times faster than \texttt{GNC}.

\subsection{Discussion}
\subsubsection{Imperfect accuracy of TLS-ARA}
From the results we can see that the accuracy of solutions to the TLS-ARA problem generally exceeds those of the TLS-RA problem, which can be explained by \emph{Remark~1}. This exhibits the superiority of angle setting to rotation matrices and ultimately results in the better robustness of \texttt{DEGNC-LAF} compared to \texttt{DEGNC-RMF}. Despite the superior accuracy of TLS-ARA to TLS-RA, the solutions to TLS-ARA can not always reach perfect accuracy (e.g., as shown in Fig. 2), which explains why we do not directly adopt the solutions to TLS-ARA as the final estimates. Fortunately, the TLS-TA problem is not too sensitive to the rotation accuracy deficiencies caused by TLS-ARA and thus can usually provide a correct set of outliers under mild rotational errors.

\subsubsection{Robustness of \texttt{DEGNC-LAF} vs. \texttt{GNC}}
We note that comparing \texttt{DEGNC-LAF} with \texttt{GNC} in terms of robostness presents opposite results on the \textsf{\small kitti\_05} and \textsf{\small manhattan} datasets. This can be explained by the fact that the odometry measurements in \textsf{\small kitti\_05} have low rotational noise but relatively high translational noise, while in \textsf{\small manhattan} the opposite is true. In scenarios like \textsf{\small kitti\_05}, \texttt{DEGNC-LAF} can obtain a good estimate of rotation by solving the first subproblem, allowing the estimate to effectively resist outliers even with high translational noise, and thus performs better robustness than \texttt{GNC}.

\section{Conclusion}
\label{sec:Conclusion}
We proposed a specialized framework for global outlier rejection over planar pose graph. Decoupling the robust PGO problem and introducing a linear representation for planar rotation are the keys to the proposed framework. This framework requires only linear solvers instead of global nonlinear solvers, for which preconditions are necessary to reach global optima and are often computationally intensive. We believe the proposed method can be an effective alternative to the standard and general-purpose \texttt{GNC} algorithm for robust planar PGO. This is especially true when the scenario allows for relatively accurate rotation estimation on odometry or when it is necessary to maintain a dense distribution of loop closures.

\newlength{\bibitemsep}\setlength{\bibitemsep}{0.00\baselineskip}
\newlength{\bibparskip}\setlength{\bibparskip}{0pt}
\let\oldthebibliography\thebibliography
\renewcommand\thebibliography[1]{
    \oldthebibliography{#1}
    \setlength{\parskip}{\bibitemsep}
    \setlength{\itemsep}{\bibparskip}
}
\bibliography{references}

\begin{thebibliography}{10}
\providecommand{\url}[1]{#1}
\csname url@rmstyle\endcsname
\providecommand{\newblock}{\relax}
\providecommand{\bibinfo}[2]{#2}
\providecommand\BIBentrySTDinterwordspacing{\spaceskip=0pt\relax}
\providecommand\BIBentryALTinterwordstretchfactor{4}
\providecommand\BIBentryALTinterwordspacing{\spaceskip=\fontdimen2\font plus
\BIBentryALTinterwordstretchfactor\fontdimen3\font minus
  \fontdimen4\font\relax}
\providecommand\BIBforeignlanguage[2]{{%
\expandafter\ifx\csname l@#1\endcsname\relax
\typeout{** WARNING: IEEEtran.bst: No hyphenation pattern has been}%
\typeout{** loaded for the language `#1'. Using the pattern for}%
\typeout{** the default language instead.}%
\else
\language=\csname l@#1\endcsname
\fi
#2}}

\bibitem{rosen2019se}
D.~M. Rosen, L.~Carlone, A.~S. Bandeira, and J.~J. Leonard, ``Se-sync: A
  certifiably correct algorithm for synchronization over the special euclidean
  group,'' \emph{The International Journal of Robotics Research}, vol.~38, no.
  2-3, pp. 95--125, 2019.

\bibitem{sunderhauf2012switchable}
N.~S{\"u}nderhauf and P.~Protzel, ``Switchable constraints for robust pose
  graph slam,'' in \emph{2012 IEEE/RSJ International Conference on Intelligent
  Robots and Systems}.\hskip 1em plus 0.5em minus 0.4em\relax IEEE, 2012, pp.
  1879--1884.

\bibitem{olson2013inference}
E.~Olson and P.~Agarwal, ``Inference on networks of mixtures for robust robot
  mapping,'' \emph{The International Journal of Robotics Research}, vol.~32,
  no.~7, pp. 826--840, 2013.

\bibitem{agarwal2013robust}
P.~Agarwal, G.~D. Tipaldi, L.~Spinello, C.~Stachniss, and W.~Burgard, ``Robust
  map optimization using dynamic covariance scaling,'' in \emph{2013 IEEE
  International Conference on Robotics and Automation}.\hskip 1em plus 0.5em
  minus 0.4em\relax Ieee, 2013, pp. 62--69.

\bibitem{latif2013robust}
Y.~Latif, C.~Cadena, and J.~Neira, ``Robust loop closing over time for pose
  graph slam,'' \emph{The International Journal of Robotics Research}, vol.~32,
  no.~14, pp. 1611--1626, 2013.

\bibitem{mangelson2018pairwise}
J.~G. Mangelson, D.~Dominic, R.~M. Eustice, and R.~Vasudevan, ``Pairwise
  consistent measurement set maximization for robust multi-robot map merging,''
  in \emph{2018 IEEE international conference on robotics and automation
  (ICRA)}.\hskip 1em plus 0.5em minus 0.4em\relax IEEE, 2018, pp. 2916--2923.

\bibitem{tzoumas2019outlier}
V.~Tzoumas, P.~Antonante, and L.~Carlone, ``Outlier-robust spatial perception:
  Hardness, general-purpose algorithms, and guarantees,'' in \emph{2019
  IEEE/RSJ International Conference on Intelligent Robots and Systems
  (IROS)}.\hskip 1em plus 0.5em minus 0.4em\relax IEEE, 2019, pp. 5383--5390.

\bibitem{yang2020graduated}
H.~Yang, P.~Antonante, V.~Tzoumas, and L.~Carlone, ``Graduated non-convexity
  for robust spatial perception: From non-minimal solvers to global outlier
  rejection,'' \emph{IEEE Robotics and Automation Letters}, vol.~5, no.~2, pp.
  1127--1134, 2020.

\bibitem{black1996unification}
M.~J. Black and A.~Rangarajan, ``On the unification of line processes, outlier
  rejection, and robust statistics with applications in early vision,''
  \emph{International journal of computer vision}, vol.~19, no.~1, pp. 57--91,
  1996.

\bibitem{chin2017maximum}
T.-J. Chin and D.~Suter, ``The maximum consensus problem: recent algorithmic
  advances,'' \emph{Synthesis Lectures on Computer Vision}, vol.~7, no.~2, pp.
  1--194, 2017.

\bibitem{sarlette2009consensus}
A.~Sarlette and R.~Sepulchre, ``Consensus optimization on manifolds,''
  \emph{SIAM journal on Control and Optimization}, vol.~48, no.~1, pp. 56--76,
  2009.

\bibitem{huber2011robust}
P.~J. Huber, ``Robust statistics,'' in \emph{International encyclopedia of
  statistical science}.\hskip 1em plus 0.5em minus 0.4em\relax Springer, 2011,
  pp. 1248--1251.

\bibitem{bosse2016robust}
M.~Bosse, G.~Agamennoni, I.~Gilitschenski, \emph{et~al.}, ``Robust estimation
  and applications in robotics,'' \emph{Foundations and Trends{\textregistered}
  in Robotics}, vol.~4, no.~4, pp. 225--269, 2016.

\bibitem{doherty2022discrete}
K.~J. Doherty, Z.~Lu, K.~Singh, and J.~J. Leonard, ``Discrete-continuous
  smoothing and mapping,'' \emph{arXiv preprint arXiv:2204.11936}, 2022.

\bibitem{Agarwal_Ceres_Solver_2022}
\BIBentryALTinterwordspacing
S.~Agarwal, K.~Mierle, and T.~C.~S. Team, ``{Ceres Solver},'' 3 2022. [Online].
  Available: \url{https://github.com/ceres-solver/ceres-solver}
\BIBentrySTDinterwordspacing

\bibitem{kummerle2011g}
R.~K{\"u}mmerle, G.~Grisetti, H.~Strasdat, K.~Konolige, and W.~Burgard, ``g 2
  o: A general framework for graph optimization,'' in \emph{2011 IEEE
  International Conference on Robotics and Automation}.\hskip 1em plus 0.5em
  minus 0.4em\relax IEEE, 2011, pp. 3607--3613.

\bibitem{hartley2003multiple}
R.~Hartley and A.~Zisserman, \emph{Multiple view geometry in computer
  vision}.\hskip 1em plus 0.5em minus 0.4em\relax Cambridge university press,
  2003.

\bibitem{antonante2021outlier}
P.~Antonante, V.~Tzoumas, H.~Yang, and L.~Carlone, ``Outlier-robust estimation:
  Hardness, minimally tuned algorithms, and applications,'' \emph{IEEE
  Transactions on Robotics}, vol.~38, no.~1, pp. 281--301, 2021.

\bibitem{chatterjee2017robust}
A.~Chatterjee and V.~M. Govindu, ``Robust relative rotation averaging,''
  \emph{IEEE transactions on pattern analysis and machine intelligence},
  vol.~40, no.~4, pp. 958--972, 2017.

\bibitem{schonberger2016structure}
J.~L. Schonberger and J.-M. Frahm, ``Structure-from-motion revisited,'' in
  \emph{Proceedings of the IEEE conference on computer vision and pattern
  recognition}, 2016, pp. 4104--4113.

\bibitem{casafranca2013back}
J.~J. Casafranca, L.~M. Paz, and P.~Pini{\'e}s, ``A back-end l 1 norm based
  solution for factor graph slam,'' in \emph{2013 IEEE/RSJ International
  Conference on Intelligent Robots and Systems}.\hskip 1em plus 0.5em minus
  0.4em\relax IEEE, 2013, pp. 17--23.

\bibitem{carlone2018convex}
L.~Carlone and G.~C. Calafiore, ``Convex relaxations for pose graph
  optimization with outliers,'' \emph{IEEE Robotics and Automation Letters},
  vol.~3, no.~2, pp. 1160--1167, 2018.

\bibitem{zhou2016fast}
Q.-Y. Zhou, J.~Park, and V.~Koltun, ``Fast global registration,'' in
  \emph{European conference on computer vision}.\hskip 1em plus 0.5em minus
  0.4em\relax Springer, 2016, pp. 766--782.

\bibitem{indelman2016incremental}
V.~Indelman, E.~Nelson, J.~Dong, N.~Michael, and F.~Dellaert, ``Incremental
  distributed inference from arbitrary poses and unknown data association:
  Using collaborating robots to establish a common reference,'' \emph{IEEE
  Control Systems Magazine}, vol.~36, no.~2, pp. 41--74, 2016.

\bibitem{karimian2020rotational}
A.~Karimian, Z.~Yang, and R.~Tron, ``Rotational outlier identification in pose
  graphs using dual decomposition,'' in \emph{European Conference on Computer
  Vision}.\hskip 1em plus 0.5em minus 0.4em\relax Springer, 2020, pp. 391--407.

\bibitem{boumal2014cramer}
N.~Boumal, A.~Singer, P.-A. Absil, and V.~D. Blondel, ``Cram{\'e}r--rao bounds
  for synchronization of rotations,'' \emph{Information and Inference: A
  Journal of the IMA}, vol.~3, no.~1, pp. 1--39, 2014.

\bibitem{carlone2014fast}
L.~Carlone, R.~Aragues, J.~A. Castellanos, and B.~Bona, ``A fast and accurate
  approximation for planar pose graph optimization,'' \emph{The International
  Journal of Robotics Research}, vol.~33, no.~7, pp. 965--987, 2014.

\bibitem{bai2021sparse}
F.~Bai, T.~Vidal-Calleja, and G.~Grisetti, ``Sparse pose graph optimization in
  cycle space,'' \emph{IEEE Transactions on Robotics}, vol.~37, no.~5, pp.
  1381--1400, 2021.

\bibitem{yang2022certifiably}
H.~Yang and L.~Carlone, ``Certifiably optimal outlier-robust geometric
  perception: Semidefinite relaxations and scalable global optimization,''
  \emph{IEEE Transactions on Pattern Analysis and Machine Intelligence}, 2022.

\bibitem{carlone2022estimation}
L.~Carlone, ``Estimation contracts for outlier-robust geometric perception,''
  \emph{arXiv preprint arXiv:2208.10521}, 2022.

\bibitem{parra2021rotation}
A.~Parra, S.-F. Chng, T.-J. Chin, A.~Eriksson, and I.~Reid, ``Rotation
  coordinate descent for fast globally optimal rotation averaging,'' in
  \emph{Proceedings of the IEEE/CVF Conference on Computer Vision and Pattern
  Recognition}, 2021, pp. 4298--4307.

\bibitem{Rosen2022Accelerating}
\BIBentryALTinterwordspacing
D.~M. Rosen, ``Accelerating certifiable estimation with preconditioned
  eigensolvers,'' May 2022. [Online]. Available:
  \url{https://arxiv.org/abs/2207.05257}
\BIBentrySTDinterwordspacing

\bibitem{gtsam}
\BIBentryALTinterwordspacing
F.~Dellaert, R.~Roberts, V.~Agrawal, A.~Cunningham, C.~Beall, D.-N. Ta,
  F.~Jiang, lucacarlone, nikai, J.~L. Blanco-Claraco, S.~Williams, ydjian,
  J.~Lambert, A.~Melim, Z.~Lv, A.~Krishnan, J.~Dong, G.~Chen, K.~Chande,
  balderdash devil, DiffDecisionTrees, S.~An, mpaluri, E.~P. Mendes, M.~Bosse,
  A.~Patel, A.~Baid, P.~Furgale, matthewbroadwaynavenio, and roderick koehle,
  ``borglab/gtsam,'' May 2022. [Online]. Available:
  \url{https://doi.org/10.5281/zenodo.5794541}
\BIBentrySTDinterwordspacing

\bibitem{Zhang18iros}
Z.~Zhang and D.~Scaramuzza, ``A tutorial on quantitative trajectory evaluation
  for visual(-inertial) odometry,'' in \emph{IEEE/RSJ Int. Conf. Intell. Robot.
  Syst. (IROS)}, 2018.

\end{thebibliography}

\end{document}